\documentclass[10pt,twocolumn,letterpaper]{article}

\usepackage{arxiv2col}

\usepackage[breaklinks,colorlinks,allcolors=linkblue]{hyperref}
\usepackage{dblfloatfix}
\usepackage{booktabs}
\usepackage{multirow}
\usepackage{xcolor}
\usepackage{pifont}
\definecolor{dgoodc}{RGB}{0,128,60}
\definecolor{dbadc}{RGB}{200,30,30}
\newcommand{\cmark}{\textcolor{green!55!black}{\ding{51}}}
\newcommand{\xmark}{\textcolor{red!75!black}{\ding{55}}}

\definecolor{prevgray}{RGB}{140,140,140}
\newcommand{\prev}[1]{\textcolor{prevgray}{#1}}

\newcommand{\dgood}[1]{\,{\textcolor{dgoodc}{(#1)}}}

\usepackage[table]{xcolor}

\usepackage{algorithm}
\usepackage{algpseudocode}
\usepackage{multirow,adjustbox}
\definecolor{oursblue}{RGB}{59,125,216}
\newcommand{\ours}[1]{\textcolor{oursblue}{#1}}
\definecolor{oursbg}{HTML}{EDF5FA}
\definecolor{gaincolor}{HTML}{007F75}

\title{Beyond Pose Control: Anatomically Consistent\\Ultrasound Video Generation}
\shorttitle{Anatomically Consistent Ultrasound Video Generation}  % running header (right)

\hypersetup{pdftitle={Beyond Pose Control: Anatomically Consistent Ultrasound Video Generation}}

\author{\name Yue Zhou \email{yue.zhou@tum.de}
\AND \name Fangxu Han \email{fangxu.han@tum.de}
\AND \name Tony Danjun Wang \email{tony.wang@tum.de}
\AND \name Nassir Navab \email{nassir.navab@tum.de}
\AND \name Yuan Bi \email{yuan.bi@tum.de}
\\[0.6ex]
\addr{Technical University of Munich, Germany}
}

\hypersetup{pdfauthor={Yue Zhou, Fangxu Han, Tony Danjun Wang, Nassir Navab, Yuan Bi}}

\begin{document}

\definecolor{tpose}{HTML}{00795A}
\definecolor{tmem}{HTML}{0072B2}
\definecolor{tprior}{HTML}{C06800}

% ===== Title block + first-page teaser (full width) =====
\twocolumn[{%
    \renewcommand\twocolumn[1][]{#1}%
    \maketitle

    \centering
    \includegraphics[
        width=\textwidth,
        height=\textheight,
        keepaspectratio
    ]{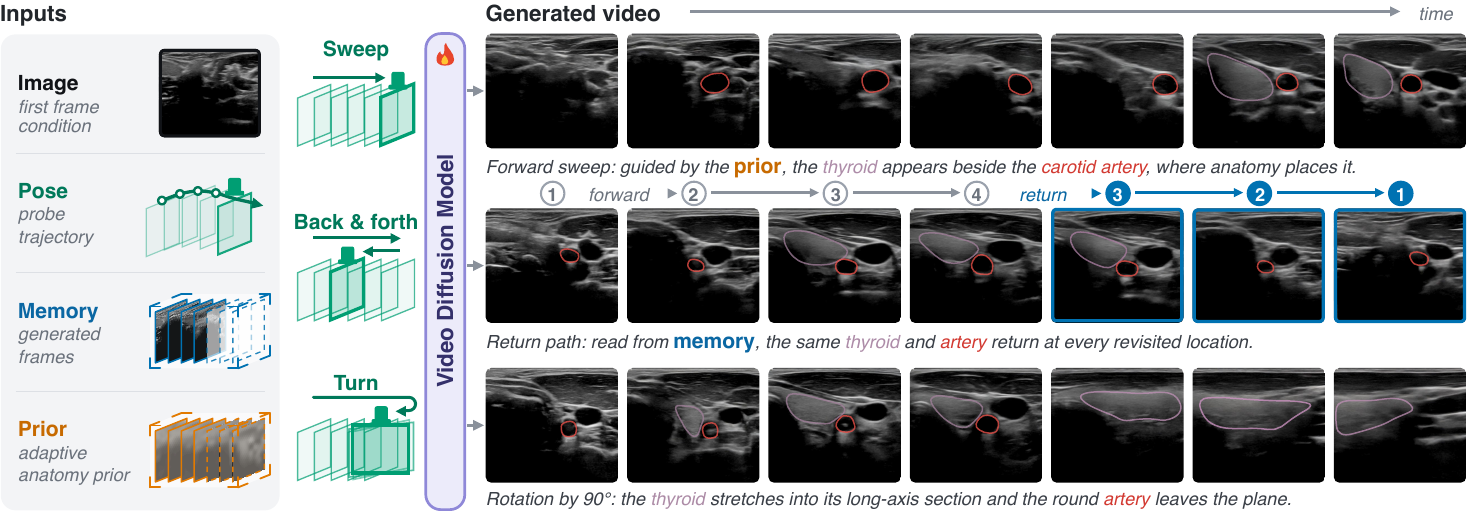}

    \captionsetup{hypcap=false}
    \captionof{figure}{
      Conditioned on a first frame, a \textcolor{tpose}{\textbf{probe trajectory}}, a 3D \textcolor{tmem}{\textbf{memory}} of generated frames, and an adaptive anatomy \textcolor{tprior}{\textbf{prior}}, our model generates plausible anatomy in unseen regions (\emph{Sweep}), stays consistent at revisited locations (\emph{Back \& forth}), and follows large rotations (\emph{Turn}). Outlines mark the thyroid and the carotid artery; highlighted frames are revisited locations.
    }
    \label{fig:teaser}

    \vspace{1.0em}
}]

\begin{abstract}
Camera-controlled video generation anticipates how a scene appears from new viewpoints by relying on a prior learned implicitly from large-scale data. 
Ultrasound poses the same problem, with the probe pose as the viewpoint, but under opposite conditions: anatomy offers an explicit prior that every sonographer relies on, whereas tracked scans are too scarce to learn it implicitly. 
As a result, pose control alone is not enough, and pose-conditioned ultrasound generators produce implausible anatomy in unscanned regions and inconsistent anatomy on revisits. We go beyond pose control and give the generator the two kinds of knowledge that make up a sonographer's mental model. An anatomical prior, built from a population of subjects, guides generation where nothing has been scanned, and is refined online by every generated chunk towards the scanned subject. A 3D latent memory, designed around the cross-sectional physics of ultrasound, compounds the generated slices into a volume of the patient and reslices it at new poses, keeping revisited anatomy consistent. The fine-grained memory enters the generator at the input, while the coarse anatomy prior guides the layout across transformer blocks. Experiments on thyroid, kidney, and lumbar spine show improved visual fidelity, pose following, and anatomical consistency over pose-conditioned and memory-augmented baselines. Code and models will be made publicly available at \url{https://karolinezhy.github.io/BeyondPoseControl/}.
\end{abstract}

\section{Introduction}
\label{sec:intro}
Humans navigate new environments with a mental prior learned from many similar ones. Entering an unfamiliar hotel room, we expect the bathroom near the door and the bed further inside; the exact layout differs from room to room, and every step refines our expectation. Video generation models~\cite{wan2025wan,hacohen2026ltx} acquire a comparable prior from large-scale data, learning how scenes typically look and move. Camera-controlled video generation~\cite{wang2024motionctrl,he2024cameractrl,bahmani2025ac3d} builds on it to anticipate how a scene appears from a new viewpoint, but this prior remains implicit in the weights. Ultrasound (US) poses the same problem, with the probe pose as the viewpoint, yet under opposite conditions. Anatomy plays the role of the room layout: the relative positions of organs are shared across people and known to every expert sonographer, while their exact shape varies between individuals and becomes clear only as the scan proceeds. 

Unlike general-domain video generation, pose-tracked ultrasound data is limited, making it difficult to implicitly learn the relationship between probe poses and 3D anatomy. Moreover, robotic ultrasound~\cite{jiang2023robotic,su2024fully,jiang2025ultrabot,11396021} requires not only accurate pose following but also consistency with the underlying 3D anatomy~\cite{bi2024machine}.

A sonographer's mental model holds two kinds of knowledge: anatomical information, which predicts what lies in unscanned tissue, and a patient-specific 3D spatial volume, which preserves what has already been scanned. Existing ultrasound world models, such as SonoWorld~\cite{yang2026sonoworld} and UltraWorld~\cite{yang2026ultraworld}, learn anatomy implicitly from expanded training data but do not explicitly maintain both forms of knowledge during generation. Natural-video generators maintain such records through retrieved frames~\cite{yu2025context,xiao2026worldmem} or 3D caches~\cite{ren2025gen3c,wang2026latent}, but these rely on camera geometry, in which neighboring views overlap and can be reprojected. 
% An ultrasound image is instead a planar cross-section, and every out-of-plane motion reveals tissue that no earlier frame contains, so a memory suited to ultrasound is still missing. 
An ultrasound image is instead a planar cross-section, and every out-of-plane motion reveals tissue that no earlier frame contains, so memory mechanisms designed for natural videos do not directly transfer to ultrasound.

Anatomically consistent ultrasound generation requires three complementary capabilities: producing plausible anatomy in unseen regions, preserving it across revisits, and responding coherently to probe motion. We therefore go beyond pose control by providing two explicit forms of anatomical knowledge. \emph{(i) What guides generation in unseen regions?} A population-level anatomical prior that, like the hotel-room expectation described earlier, is progressively refined by generated observations to adapt to the subject. \emph{(ii) What preserves anatomy across revisits?} A 3D latent memory that compounds generated slices and reslices them at new probe poses, following ultrasound imaging geometry rather than camera projection. Together, the prior guides unseen anatomy while the memory preserves previously generated structures.

The two conditions play distinct roles: the memory provides fine-grained content at known locations, while the prior offers a coarse estimate of anatomical layout. We therefore inject them differently: memory is concatenated with the video latent at the input, whereas the prior guides early transformer blocks through residual conditioning. Our contributions are summarized as follows:
\begin{itemize}
\item We propose an explicit anatomical prior for ultrasound video generation: a population model of anatomy that is updated online from the generated frames, so that it adapts the prior to the subject even in regions not yet scanned.
\item We propose a compounding-based 3D latent memory, the first memory for video generation designed around cross-sectional imaging rather than camera geometry, which keeps revisited regions consistent.
\item We propose a dual-path conditioning: the fine-grained memory is concatenated with the input, while the coarse prior is injected across transformer blocks to guide the layout without being treated as fine detail.
\end{itemize}
Experiments on thyroid, kidney, and lumbar spine show that our framework outperforms existing models in fidelity, pose following, revisit consistency, and expert rating.

\section{Related Works}
\label{sec:relatedworks}

\paragraph{Video generation.}
Diffusion models have progressed from latent image synthesis~\cite{rombach2022high} to video generation with spatiotemporal U-Nets~\cite{ho2022video,blattmann2023align}, and large-scale video diffusion transformers such as Wan~\cite{wan2025wan}, LTX-2~\cite{hacohen2026ltx} and MiniMax-H3~\cite{minimaxai2026h3_hf} now achieve high visual fidelity and motion realism. Controllable generation extends them with camera signals~\cite{bai2025recammaster}: MotionCtrl~\cite{wang2024motionctrl} disentangles camera and object motion, CameraCtrl~\cite{he2024cameractrl} encodes camera geometry for trajectory-guided generation, and AC3D~\cite{bahmani2025ac3d} studies how pose conditioning should enter video diffusion models. In all of them, what appears at a new viewpoint is determined by a prior learned implicitly from large-scale data.

\paragraph{Spatially consistent video generation.}
DFoT~\cite{song2025history, chen2024diffusion} conditions on generation history to keep autoregressive segments continuous, and memory mechanisms extend consistency to longer horizons. Context-as-Memory~\cite{yu2025context} retrieves historical frames by camera-view overlap, and WorldMem~\cite{xiao2026worldmem} reads a memory bank through state-aware attention. GEN3C~\cite{ren2025gen3c}, Spatia~\cite{zhao2026spatia} and Voyager~\cite{huang2025voyager} maintain explicit point-cloud caches that are rendered at new viewpoints, while latent spatial memory~\cite{wang2026latent} stores features directly in the diffusion latent space. We share the idea of a persistent 3D memory with these works, but all of them build it through camera projection and depth-based unprojection. For ultrasound, we replace this with compounding, the standard way of assembling tracked cross-sections into a volume~\cite{SOLBERG2007991}.

\paragraph{Ultrasound world models.}
Robotic ultrasound has motivated world models~\cite{ha2018world} that relate probe motion to anatomical observations. SonoGym~\cite{ao2026sonogym} provides a physics-based simulation environment for policy learning, and Cardiac Copilot~\cite{jiang2024cardiac} and EchoWorld~\cite{yue2025echoworld} learn latent representations of neighboring planes and motion-induced dynamics for probe guidance~\cite{kang2026dreamreg,bi2022vesnet}. Generative world models predict ultrasound frames directly. Fan~\etal~\cite{fan2026action} condition a diffusion model on tracked probe motion for goal-plane navigation. SonoWorld~\cite{yang2026sonoworld} pretrains on millions of untracked clinical frames, grounds the learned dynamics to EM-tracked probe poses, and generates causally with a KV cache. UltraWorld~\cite{yang2026ultraworld} synthesizes action--video pairs from 3D anatomical masks and conditions on an acoustic sampling map of per-pixel 3D sampling geometry, similar in spirit to our pose encoding. All of these condition on the probe pose and learn anatomy implicitly; our method additionally provides an explicit updated anatomical prior and a spatial memory of the scanned patient.

\begin{algorithm}[t]
\caption{Chunk-wise generation. Black: pose-conditioned models; \prev{gray}: memory-augmented models; \ours{blue}: ours for ultrasound. $\mathcal{A}$: anatomical prior; $\mathcal{M}$: memory volume.}
\label{alg:rollout}
\small
\begin{algorithmic}[1]
\Require first frame $\mathbf{I}_0$, poses $\mathbf{P}^{(1:K)}$ of $K$ chunks
\State $\mathbf{X}_{\mathrm{ctx}} \gets \mathbf{I}_0$
\State \ours{$\mathcal{A} \gets \boldsymbol{\mu}$} \Comment{Sec.~\ref{sec:prior}}
\State \prev{$\mathcal{M} \gets$} \ours{empty volume} \Comment{Sec.~\ref{sec:memory}}
\For{$k = 1,\dots,K$}
  \State \ours{$\mathbf{c}_{\mathcal{P}} \gets \textsc{PointMap}(\mathbf{P}^{(k)})$} \Comment{Eq.~\ref{eq:pose_token}}
  \State \prev{$\mathbf{c}_{\mathcal{M}} \gets$} \ours{$\textsc{Reslice}(\mathcal{M}, \mathbf{P}^{(k)})$} \Comment{Eq.~\ref{eq:reslicing}}
  \State \ours{$\mathbf{c}_{\mathcal{A}} \gets \textsc{Reslice}(\mathcal{A}, \mathbf{P}^{(k)})$} \Comment{Eq.~\ref{eq:reslicing}}
  \State $\hat{\mathbf{X}}^{(k)} \sim p_\theta\big(\mathbf{X}^{(k)} \mid \mathbf{X}_{\mathrm{ctx}}, \ours{\mathbf{c}_{\mathcal{P}}}, \prev{\mathbf{c}_{\mathcal{M}}}, \ours{\mathbf{c}_{\mathcal{A}}}\big)$ \Comment{Eq.~\ref{eq:concat}}
  \State \prev{$\mathcal{M} \gets$} \ours{$\textsc{Compound}(\mathcal{M}, \hat{\mathbf{X}}^{(k)}, \mathbf{P}^{(k)})$} \Comment{Eq.~\ref{eq:memory_update}}
  \State \ours{$\mathcal{A} \gets \textsc{Update}(\mathcal{M})$} \Comment{Eq.~\ref{eq:prior_update}}
  \State $\mathbf{X}_{\mathrm{ctx}} \gets$ last frame of $\hat{\mathbf{X}}^{(k)}$
\EndFor
\State \Return $\hat{\mathbf{X}}^{(1:K)}$
\end{algorithmic}
\end{algorithm}

% ------------------------------------------------------------
\section{Method}
\label{sec:method}

\begin{figure*}
    \centering
    \includegraphics[width=\linewidth]{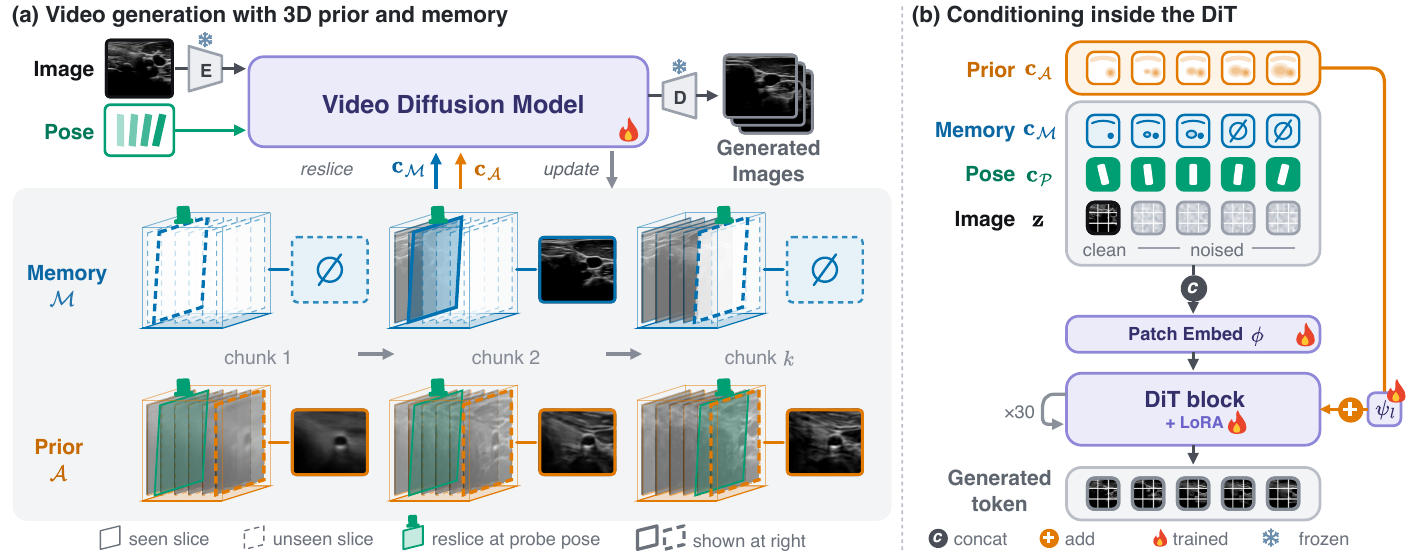}
\caption{\textbf{Method overview.}
(a) Each chunk is conditioned on reslices of the 3D \textcolor{tmem}{memory $\mathcal{M}$} and \textcolor{tprior}{prior $\mathcal{A}$} at its target \textcolor{tpose}{poses}; the generated frames then update both volumes. A \textcolor{tmem}{memory} reslice may be empty ($\varnothing$) or partially filled, with a mask marking where it is valid; the \textcolor{tprior}{prior} is defined everywhere.
(b) \textcolor{tmem}{Memory} and \textcolor{tpose}{pose} are concatenated with the video latent and patch-embedded; the \textcolor{tprior}{prior} is added as a residual.}

\label{fig:method}
\end{figure*}

% ------------------------------------------------------------
\subsection{Problem Statement}
\label{sec:problem}

We consider pose-conditioned ultrasound video generation given
an initial frame and a probe trajectory expressed in a shared
anatomical coordinate system. Following existing pose-conditioned
video generation models~\cite{he2024cameractrl,song2025history,wang2024motionctrl},
we generate long videos chunk by chunk using a conditional
generative model $p_\theta$. Each chunk $\mathbf{X}^{(k)}$
is conditioned on its probe poses $\mathbf{P}^{(k)}$ and
the context $\mathbf{X}_{\mathrm{ctx}}$, the last frame
of the previous chunk:
\begin{equation}
\hat{\mathbf{X}}^{(k)} \sim p_{\theta}\big(
\mathbf{X}^{(k)} \mid
\mathbf{X}_{\mathrm{ctx}}, \mathbf{P}^{(k)}
\big).
\label{eq:pose_cond}
\end{equation}
However, this limited context does not explicitly preserve
earlier generated content. Memory-augmented
models~\cite{xiao2026worldmem,yu2025context,ren2025gen3c}
therefore maintain a memory $\mathcal{M}$ of previous
generations. Before each chunk, they retrieve the memory
condition $\mathbf{c}_{\mathcal{M}} =
\mathcal{R}(\mathcal{M},\mathbf{P}^{(k)})$
at the target poses, and afterward update the memory with
the generated chunk,
$\mathcal{M}\gets\textsc{Update}
(\mathcal{M},\hat{\mathbf{X}}^{(k)},\mathbf{P}^{(k)})$.

Ultrasound scanning requires both plausible anatomy in unscanned regions and consistent anatomy in revisited ones. We therefore condition each chunk on an anatomical prior $\mathcal{A}$ (Sec.~\ref{sec:prior}) alongside the memory $\mathcal{M}$ (Sec.~\ref{sec:memory}). The prior is read in the same way as the memory, $\mathbf{c}_{\mathcal{A}}=\mathcal{R}(\mathcal{A},\mathbf{P}^{(k)})$, and is updated from the memory after each chunk, $\mathcal{A}\gets\textsc{Update}(\mathcal{M})$. Each chunk is then generated as
\begin{equation}
\hat{\mathbf{X}}^{(k)} \sim p_{\theta}\big(\mathbf{X}^{(k)} \mid \mathbf{X}_{\mathrm{ctx}}, 
\mathbf{c}_{\mathcal{P}}, 
\mathbf{c}_{\mathcal{M}}, \mathbf{c}_{\mathcal{A}}\big).
\label{eq:problem}
\end{equation}
Alg.~\ref{alg:rollout} summarizes the procedure. The following sections define $\mathcal{A}$, $\mathcal{M}$, their updates for ultrasound. There we describe a single chunk and omit the chunk index.

% ------------------------------------------------------------
\subsection{Prior and Memory Construction}
\label{sec:prior_memory}

The anatomical prior $\mathcal{A}$ and the memory $\mathcal{M}$ are two latent volumes in $\mathbb{R}^{C\times X\times Y\times Z}$, defined in an anatomical coordinate system shared by all scans of an organ (Sec.~\ref{sec:experiment}).
The prior estimates plausible anatomy everywhere, including unobserved regions, while the memory stores the latent features generated so far. Both are updated after each chunk and read at the target poses by reslicing (Sec.~\ref{sec:reslicing}).

\subsubsection{Prior Construction}
\label{sec:prior}

\textit{Initialization.}
The prior is built from the training subjects. The latent frames of each subject are compounded into a volume $\mathbf{V}$ at their poses, and a low-rank Gaussian model \cite{albrecht2013posterior} is fitted across subjects:
\begin{equation}
\mathbf{V} = \boldsymbol{\mu} + \mathbf{B}\boldsymbol{\alpha} + \boldsymbol{\epsilon},
\label{eq:ppca}
\end{equation}
where $\boldsymbol{\mu}$ is the population mean, $\mathbf{B}$ contains the principal modes of anatomical variation with coefficients $\boldsymbol{\alpha}\sim\mathcal{N}(\mathbf{0},\boldsymbol{\Lambda})$, and $\boldsymbol{\epsilon}$ is residual noise. For each new generation, the prior starts from the population mean, $\mathcal{A}=\boldsymbol{\mu}$.

\noindent
\textit{Update.}
During generation, the evidence about the subject is what the model has generated so far, and this is exactly what the memory stores (Sec.~\ref{sec:memory}). After each chunk is generated and written into the memory, the prior is adapted to the subject by Bayesian conditioning on the generated content:
\begin{equation}
\mathcal{A} \leftarrow \boldsymbol{\mu}+\mathbf{B}\hat{\boldsymbol{\alpha}},
\label{eq:prior_update}
\end{equation}
where $\hat{\boldsymbol{\alpha}}$ is the posterior mean of the mode coefficients given the memory. The expectation is thus refined by what has been generated: since the principal modes in $\mathbf{B}$ capture how anatomy varies jointly across subjects, the scanned slices also inform regions not yet scanned, so the guidance there becomes increasingly specific to the subject. 
% The derivation of the update is given in the supplementary material.

% ------------------------------------------------------------
\subsubsection{Memory Construction}
\label{sec:memory}

\textit{Initialization.}
Following ultrasound compounding, the memory keeps a feature sum $\mathbf{F}$ and an observation count $\mathbf{W}$ per voxel, and stores their ratio, the average feature. Unlike the prior, it starts empty, with $\mathbf{F}=\mathbf{W}=\mathbf{0}$.

\noindent
\textit{Update.}
After each chunk, its generated latent frames are compounded into the memory at their poses. For a frame with pose $\mathbf{P}$, which maps an image location $(u,v)$ to its 3D position $\mathbf{P}(u,v)$ in the anatomical coordinate system, the latent feature $\mathbf{z}(u,v)$ is assigned to the nearest voxel $\boldsymbol{\xi}$, and
\begin{equation}
\mathbf{F}(\boldsymbol{\xi})\leftarrow\mathbf{F}(\boldsymbol{\xi})+\mathbf{z}(u,v),\qquad
\mathbf{W}(\boldsymbol{\xi})\leftarrow\mathbf{W}(\boldsymbol{\xi})+1,
\label{eq:memory_update}
\end{equation}
with $\mathcal{M}=\mathbf{F}/\mathbf{W}$ on the voxels with $\mathbf{W}>0$. This is standard pixel-based compounding, performed in latent space, so revisited regions are averaged rather than overwritten.
% ------------------------------------------------------------
\subsubsection{Pose-Conditioned Reslicing}
\label{sec:reslicing}

To condition a chunk, both volumes are resliced at its target poses $\mathbf{P}$:
\begin{equation}
\mathbf{c}_{\mathcal{A}}(u,v)=\mathcal{R}\big(\mathcal{A},\mathbf{P}(u,v)\big),
\mathbf{c}_{\mathcal{M}}(u,v)=\mathcal{R}\big(\mathcal{M},\mathbf{P}(u,v)\big),
\label{eq:reslicing}
\end{equation}
where $\mathcal{R}$ here for ultrasound is a reslicing operator that samples $\mathcal{A}$ or $\mathcal{M}$ at the positions $\mathbf{P}(u,v)$ of all image locations $(u,v)$ of the target frames. The prior condition $\mathbf{c}_{\mathcal{A}}$ provides the expected anatomy at each target plane, and the memory condition $\mathbf{c}_{\mathcal{M}}$ provides the content generated there before.

% ------------------------------------------------------------
\subsection{Model Architecture}
\label{sec:arch}
The pose, the memory and the prior differ in how reliably they align with the target frame, so we inject them differently into a video diffusion transformer as shown in  Fig.~\ref{fig:method}.

\paragraph{Pose embedding.}
Camera-controlled models encode the pose by per-pixel viewing rays, \eg, Plücker coordinates~\cite{he2024cameractrl}, since a camera pixel does not know the depth of what it shows. An ultrasound image is instead a cross-section: each pixel records the echo from a single point of tissue at a known depth along the beam, so its 3D position can be determined by the probe pose. Similar to UltraWorld~\cite{yang2026ultraworld}, we therefore embed the pose densely, as the 3D position of every image location in the anatomical coordinate system:
\begin{equation}
\mathbf{c}_{\mathcal{P}}(u,v)=\tfrac{1}{\ell}\,\mathbf{P}(u,v),
\label{eq:pose_token}
\end{equation}
where $\mathbf{P}(u,v)$ is the 3D position of image location $(u,v)$ (Sec.~\ref{sec:memory}) and $\ell$ is a scaling factor. The pose embedding is thus turned from a 6-DoF vector into a dense condition that gives every token its own 3D position.

% ------------------------------------------------------------

\paragraph{Pose and memory conditions.}
The pose and the memory provide fine-grained information for each token: its exact position and the content generated there before. We therefore concatenate them channel-wise with the image latent and patch-embed them together (Fig.~\ref{fig:method}b),
\begin{equation}
\mathbf{h}_0=\phi\big(\big[\,\mathbf{z} \,\|\, \mathbf{c}_{\mathcal{P}} \,\|\, \mathbf{c}_{\mathcal{M}}\,\big]\big),
\label{eq:concat}
\end{equation}
where $\phi$ is the pretrained patch embedding, extended with zero-initialized weights for the new channels, and $\mathbf{h}_0$ is the input to the first transformer block. In $\mathbf{z}$, the first frame is the clean context and the remaining frames are noised. Memory locations that have not been observed, whether a whole frame ($\varnothing$ in Fig.~\ref{fig:method}) or only part of it, are set to zero and marked by an extra mask channel.

\paragraph{Prior condition.}
Concatenating the prior in the same way would present it as fine detail, alongside the memory, although it is only a coarse estimate. We therefore keep it in a separate path and inject it repeatedly, so that it continues to guide the layout: $\mathbf{c}_{\mathcal{A}}$ is projected onto the token grid and added as a residual to the input of each of the first $L$ transformer blocks,
\begin{equation}
\mathbf{h}_l \leftarrow \mathbf{h}_l + \psi_l\big(\mathbf{c}_{\mathcal{A}}\big),\qquad l<L,
\label{eq:prior_inj}
\end{equation}
where $\mathbf{h}_l$ is the input of block $l$, and $\psi_l$ is a zero-initialized projection. We inject it only into the early blocks, since the prior is meant as a coarse layout cue and early blocks of video diffusion transformers mainly shape the global layout~\cite{bahmani2025ac3d}.

\subsection{Training and Inference}
\label{sec:train}

\noindent\textit{Training.}
Each training sample consists of a trajectory and a target chunk. The ground-truth latents of the preceding chunks are compounded into the memory (Eq.~\ref{eq:memory_update}), from which the prior is updated (Eq.~\ref{eq:prior_update}); to avoid leaking the target anatomy, the prior is built without the subject of the sample. Both volumes are then resliced at the target poses (Eq.~\ref{eq:reslicing}), which together with the pose condition gives $\mathbf{c}=\{\mathbf{c}_{\mathcal{P}},\mathbf{c}_{\mathcal{M}},\mathbf{c}_{\mathcal{A}}\}$.

The generator $p_\theta$ is a rectified-flow model with velocity network $v_\theta$. Given the clean latent $\mathbf{z}_0$ of the target chunk and Gaussian noise $\boldsymbol{\epsilon}\sim\mathcal{N}(\mathbf{0},\mathbf{I})$, the noisy latent and the target velocity are
\begin{equation}
\mathbf{z}_t=(1-t)\,\mathbf{z}_0+t\,\boldsymbol{\epsilon},\qquad
\mathbf{v}^{\star}=\boldsymbol{\epsilon}-\mathbf{z}_0,
\label{eq:flow_path}
\end{equation}
where $t\in[0,1]$ is sampled from the timestep distribution of the backbone. The network input $\mathbf{z}$ of Eq.~\eqref{eq:concat} consists of the clean context frame followed by $\mathbf{z}_t$, and $v_\theta$ is trained to predict the target velocity,
\begin{equation}
\mathcal{L}=
\mathbb{E}_{t,\mathbf{z}_0,\boldsymbol{\epsilon}}
\big\|v_\theta(\mathbf{z},t\mid\mathbf{c})-\mathbf{v}^{\star}\big\|_2^2 .
\label{eq:fm}
\end{equation}
We train only the new channels of $\phi$, the prior projections $\psi_l$, and LoRA adapters in the transformer blocks; all other weights are frozen (Fig.~\ref{fig:method}).

\noindent\textit{Inference.}
Generation follows Alg.~\ref{alg:rollout}: each chunk is sampled with the backbone's ODE sampler and compounded into the memory, from which the prior is updated. The only inputs are the first frame and the probe poses.
\section{Experiment}
\label{sec:experiment}

\subsection{Experiment Setup}
We collect freehand ultrasound scans of the thyroid, kidney, and lumbar spine from volunteers under an ethics-approved protocol, with probe poses tracked by an NDI system at about 30 frames per second. All recordings are mapped to a shared anatomical coordinate system through coarse landmark-based alignment, with the origin at a standard landmark of each organ, followed by a subject-level training/test split (thyroid 83/10, kidney 61/23, spine 96/21 recordings).

We study pose-conditioned video generation under this aligned-pose setting, with the initial frame's anatomical pose estimated through landmark-based alignment. Since all later poses are relative to the first frame, only a coarse placement of the scan is required.

\paragraph{Training data.}
From each scan we sample three types of training trajectories, each emphasizing one source of information. \emph{Sweeps} advance into unscanned tissue, where the model must rely on the prior. \emph{Returns} revisit scanned regions, where it must rely on the memory. \emph{Turns} rotate the probe between transverse and longitudinal views, where unseen planes must be inferred from both.

\paragraph{Baselines.}
We compare with four pose-conditioned video models that span the design space of memory: CameraCtrl~\cite{he2024cameractrl} keeps no memory; WorldMem~\cite{xiao2026worldmem} retrieves past frames by pose and fuses them through memory attention; CaM~\cite{yu2025context} retrieves co-visible frames and appends them to the context; Gen3C~\cite{ren2025gen3c} maintains a pixel-space 3D cache. For a fair comparison, all baselines are retrained on our data with the same input, and every memory holds the method's own generations. Two components are adapted from natural images to ultrasound. (i)~\emph{Pose embedding}: per-pixel viewing rays are replaced by the per-pixel 3D coordinates of the imaging plane, as in ours. (ii)~\emph{Memory storage}: frames are placed in 3D by their known plane geometry (compounding) instead of depth-based unprojection, and retrieval follows plane overlap. 
% Per-method adaptations and backbones are described in the supplementary material.

\paragraph{Evaluation metrics.}
% We evaluate anatomy consistency on each organ from three aspects. (i)~\emph{Fidelity} measures realism: FID and FVD against real scans, and LPIPS to the ground truth on regions scanned for the first time. (ii)~\emph{Revisit consistency} measures whether the model remembers: the probe returns over the scanned region with a tilt of up to $\pm11^\circ$, so no earlier frame can simply be copied, and each revisited frame is compared in PSNR and LPIPS with the real outbound scan, compounded into a volume and resliced at the tilted pose. (iii)~\emph{Pose following} measures whether generation obeys the probe: the ratio of observed to commanded displacement of an organ-specific landmark under $\pm8$\,mm lateral shifts. (iv)~\emph{Realism score} is assessed in a blinded user study, in which $N$~raters with ultrasound experience score each method's videos on a 1--5 scale. (Sc.).

We evaluate anatomical consistency on each organ from three complementary aspects.
(i)~\emph{Fidelity} measures whether generated anatomy matches real scans, using FID and FVD for realism and LPIPS against ground truth in newly scanned regions.
(ii)~\emph{Revisit consistency} measures whether the same anatomy is preserved when the probe returns over a scanned region with a slight rotation: each revisited frame is compared in PSNR and LPIPS with the real scan.
(iii)~\emph{Pose following} measures whether anatomical structures change consistently with probe motion, using the ratio of observed to commanded displacement of an organ-specific landmark under certain lateral shifts.
Beyond these metrics, we assess perceived realism in a blinded \emph{expert rating}, in which sonographers score each method's videos on a 1--5 scale (Sc.).

\begin{table*}[t]
\centering
\caption{\textbf{Quantitative comparison} on three organs in terms of fidelity, revisit consistency, pose following and expert score. Best in \textbf{bold}, second best \underline{underlined}.}
\label{tab:baseline}
\footnotesize
\setlength{\tabcolsep}{1.6pt}
\renewcommand{\arraystretch}{1.12}
\begin{adjustbox}{max width=\textwidth}
\begin{tabular}{@{}l cccccc @{\hspace{5pt}} cccccc @{\hspace{5pt}} cccccc @{\hspace{5pt}} c@{}}
\toprule
& \multicolumn{6}{c}{\textbf{Thyroid}} & \multicolumn{6}{c}{\textbf{Kidney}} & \multicolumn{6}{c}{\textbf{Spine}} & \\
\cmidrule(lr){2-7}\cmidrule(lr){8-13}\cmidrule(lr){14-19}
& \multicolumn{3}{c}{Fidelity} & \multicolumn{2}{c}{Revisit} & Pose
& \multicolumn{3}{c}{Fidelity} & \multicolumn{2}{c}{Revisit} & Pose
& \multicolumn{3}{c}{Fidelity} & \multicolumn{2}{c}{Revisit} & Pose
& \multirow{2}{*}{Sc.$\uparrow$} \\
\cmidrule(lr){2-4}\cmidrule(lr){5-6}\cmidrule(lr){7-7}\cmidrule(lr){8-10}\cmidrule(lr){11-12}\cmidrule(lr){13-13}\cmidrule(lr){14-16}\cmidrule(lr){17-18}\cmidrule(lr){19-19}
\textbf{Method}
& FID$\downarrow$ & FVD$\downarrow$ & LPIPS$\downarrow$ & PSNR$\uparrow$ & LPIPS$\downarrow$ & Fol.$\uparrow$
& FID$\downarrow$ & FVD$\downarrow$ & LPIPS$\downarrow$ & PSNR$\uparrow$ & LPIPS$\downarrow$ & Fol.$\uparrow$
& FID$\downarrow$ & FVD$\downarrow$ & LPIPS$\downarrow$ & PSNR$\uparrow$ & LPIPS$\downarrow$ & Fol.$\uparrow$
& \\
\midrule
CameraCtrl
& 208 & 675 & 0.39 & 16.41 & 0.44 & \underline{0.49}
& 170 & 504 & 0.33 & 13.83 & 0.44 & \underline{0.25}
& 163 & 674 & 0.34 & 17.79 & 0.40 & \underline{0.36}
& 2.01 \\
WorldMem
& 136 & 436 & \underline{0.35} & 15.99 & 0.47 & 0.22
& 312 & 678 & 0.37 & 14.66 & 0.49 & 0.00
& 218 & 916 & 0.35 & 15.66 & 0.43 & 0.00
& 1.54 \\
CaM
& \underline{34} & 184 & \underline{0.35} & \underline{17.10} & \underline{0.37} & 0.18
& \underline{46} & 229 & \underline{0.28} & 17.82 & 0.29 & 0.03
& 65 & 294 & 0.29 & 19.05 & 0.35 & 0.09
& 3.73 \\
Gen3C
& 38 & \underline{119} & 0.36 & 16.80 & 0.40 & 0.08
& 116 & \textbf{181} & \underline{0.28} & \underline{18.24} & \underline{0.28} & 0.00
& \underline{57} & \underline{177} & \textbf{0.28} & \textbf{19.93} & \textbf{0.31} & 0.02
& \underline{3.89} \\
\midrule
\textbf{Ours}
& \textbf{30} & \textbf{107} & \textbf{0.32} & \textbf{17.42} & \textbf{0.36} & \textbf{0.73}
& \textbf{38} & \underline{197} & \textbf{0.25} & \textbf{18.71} & \textbf{0.26} & \textbf{0.68}
& \textbf{45} & \textbf{176} & \textbf{0.28} & \underline{19.79} & \textbf{0.31} & \textbf{0.43}
& \textbf{4.25} \\
\bottomrule
\end{tabular}
\end{adjustbox}
\end{table*}

\subsection{Performance Against Baselines}
\paragraph{Fidelity.}
We measure image-level and temporal realism with FID and FVD, and accuracy on regions seen for the first time with LPIPS (see Fidelity in Tab.~\ref{tab:baseline}).
Our method obtains the lowest FID on all three organs and the lowest FVD on Thyroid and Spine, trailing only Gen3C~\cite{ren2025gen3c} in FVD on Kidney.
Its LPIPS on new regions is the lowest on Thyroid and Kidney and ties with Gen3C on Spine.
CameraCtrl~\cite{he2024cameractrl} and WorldMem~\cite{xiao2026worldmem} produce far less realistic videos, which we attribute mainly to their backbones: a UNet-based diffusion model~\cite{he2024cameractrl} pretrained on natural videos that transfers poorly to ultrasound, and a much smaller Diffusion Transformer with limited capacity for fine ultrasound texture~\cite{xiao2026worldmem}.
CaM and Gen3C, re-implemented on our Wan2.2-5B backbone and fine-tuned on ultrasound, generally reach competitive fidelity.

\paragraph{Revisit.}
When the probe returns over a scanned region along a tilted path, we compare each generated frame with the real scan (see Revisit in Tab.~\ref{tab:baseline}).
CameraCtrl~\cite{he2024cameractrl} has no memory and therefore cannot recall earlier content.
WorldMem~\cite{xiao2026worldmem} shows that memory alone is not sufficient: feeding a few retrieved past frames through attention remains a weak constraint, and its revisit scores are no better than those of CameraCtrl on Thyroid and Spine.
CaM~\cite{yu2025context} and Gen3C~\cite{ren2025gen3c}, with memory adapted to ultrasound geometry, are the closest competitors, yet our 3D latent memory is best on Thyroid and Kidney in both PSNR and LPIPS.
On Spine it ties with Gen3C in LPIPS and is within 0.14\,dB in PSNR.

\begin{figure}[b]
  \centering
  \includegraphics[width=\linewidth]{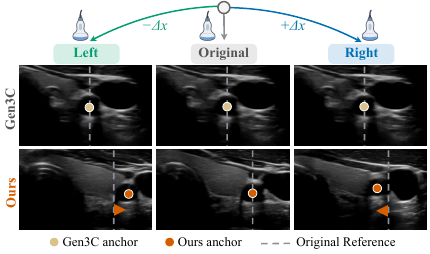}
  \caption{\textbf{Pose following.} Each method generates the original sweep and the same sweep shifted by $\Delta x=8$\,mm. Our anatomy (dot) moves with the probe; Gen3C stays at the original position.}
  \label{fig:posedev_follow}
\end{figure}

\paragraph{Pose following.}
This metric asks whether the generated anatomy actually follows the scan path: we shift the path laterally by $\pm\Delta x$ and measure, how far the target anatomy moves relative to the commanded shift (1 = perfect following).
None of the baselines uses an anatomical prior, and all of them follow the path poorly.
WorldMem, CaM, and Gen3C almost fail (at most 0.22), and CameraCtrl, the best baseline, reaches only 0.25--0.49.
This suggests that a pose signal alone is too weak a condition: the model learns a plausible motion prior but not how the anatomy should change with the probe position.
Fig.~\ref{fig:posedev_follow} shows a typical thyroid case.
When the probe moves left or right, our carotid artery moves in the opposite direction in the image, as it should, by roughly 6\,mm for an 8\,mm probe shift; Gen3C keeps it at the same position in all three sweeps despite its explicit 3D cache.
With the anatomical prior, our method improves following markedly on every organ (0.73, 0.68, and 0.43 on thyroid, kidney, and spine), consistent with the ablation in Tab.~\ref{tab:abl_mem_prior}.

\paragraph{Expert rating.}
Five sonographers each rated all 180 videos: 30 test cases (ten per organ, covering back-and-forth sweeps, probe turns and short sweeps), each shown as the real recording and the generations of the five methods. Videos are shown one at a time in random order without method names, and each is scored from 1 (definitely generated) to 5 (definitely real), so the score reflects whether a whole clip, including its consistency as the probe moves, passes as a real scan. Our videos receive the highest score among all generators (4.25, versus 3.89 for Gen3C and 3.73 for CaM; Tab.~\ref{tab:baseline}), approaching the real recordings (4.49). Details are given in the supplementary material.

% \begin{figure*}[t]
%   \centering
%   \begin{minipage}[t]{0.49\textwidth}
%     \centering
%     \includegraphics[width=\linewidth]{images/fig_posedev_follow.pdf}
%     \caption{\textbf{Pose following.} Each method generates the original sweep and the same sweep shifted by $\Delta x=8$\,mm. Our anatomy (dot) moves with the probe; Gen3C stays at the original position.}
%     \label{fig:posedev_follow}
%   \end{minipage}\hfill
%   \begin{minipage}[t]{0.49\textwidth}
%     \centering
%     \includegraphics[width=\linewidth]{images/prior.pdf}
%     \caption{\textbf{Static vs.\ dynamic prior.} The dynamic prior lowers the prior error (distance to the real anatomy at unscanned planes) and the FID (left), and improves pose following and revisit PSNR (right).}
% \label{fig:prior_update}
%   \end{minipage}
% \end{figure*}

\begin{table*}[t]
\centering
% ---------- 左:Memory / Prior 消融 ----------
\begin{minipage}[t]{0.43\textwidth}
\centering
\caption{\textbf{Ablation on the components and their injection.}}
\label{tab:abl_mem_prior}
\setlength{\tabcolsep}{4pt}
\renewcommand{\arraystretch}{1.12}
\resizebox{\linewidth}{!}{
\begin{tabular}{cc c cc c}
\toprule
\multirow{2}{*}{\textbf{Memory}} & \multirow{2}{*}{\textbf{Prior}}
& \textbf{Fidelity} & \multicolumn{2}{c}{\textbf{Revisit PSNR}$\uparrow$} & \textbf{Pose} \\
\cmidrule(lr){3-3}\cmidrule(lr){4-5}\cmidrule(lr){6-6}
& & FID$\downarrow$ & GT & Self & Fol.$\uparrow$ \\
\midrule
\multicolumn{6}{l}{\emph{Whether to use each condition}} \\
\xmark & \xmark & 40.2 & 16.85 & 18.32 & 0.10 \\
\xmark & \cmark & 30.7 & 17.22 & 18.59 & 0.58 \\
\cmark & \xmark & 31.2 & 16.94 & 27.74 & 0.01 \\
\midrule
\multicolumn{6}{l}{\emph{How to inject each condition}} \\
Layer-wise & Layer-wise & 37.9 & 17.11 & 28.38 & 0.53 \\
Concat & Concat & 33.2 & 17.18 & 27.71 & 0.58 \\
\midrule
\rowcolor{gray!12}
Concat & Layer-wise & \textbf{30.4} & \textbf{17.42} & \textbf{28.40} & \textbf{0.73} \\
\bottomrule
\end{tabular}
}
\end{minipage}
\hfill
% ---------- 右:换底座 ----------
\begin{minipage}[t]{0.53\textwidth}
\centering

\caption{\textbf{Model-agnostic improvement.} Our memory and prior consistently improve video diffusion backbones of different architectures and sizes\dgood{change from the base model}.}

\label{tab:model_agnostic}
\setlength{\tabcolsep}{4pt}
\renewcommand{\arraystretch}{1.12}
\resizebox{\linewidth}{!}{
\begin{tabular}{l c cc c}
\toprule
& \textbf{Fidelity} & \multicolumn{2}{c}{\textbf{Revisit PSNR}$\uparrow$} & \textbf{Pose} \\
\cmidrule(lr){2-2}\cmidrule(lr){3-4}\cmidrule(lr){5-5}
\textbf{Method} & FVD$\downarrow$ & GT & Self & Fol.$\uparrow$ \\
\midrule
Wan2.2-5B~\cite{wan2025wan} & 132 & 16.9 & 18.3 & 0.10 \\
\rowcolor{gray!12}
\quad + Ours & \textbf{107}\dgood{$-$19\%} & \textbf{17.4}\dgood{+3\%} & \textbf{28.4}\dgood{+55\%} & \textbf{0.73}\dgood{+0.63} \\
\midrule
LTX2.5-22B~\cite{hacohen2026ltx} & 151 & 15.6 & 17.5 & 0.04 \\
\rowcolor{gray!12}
\quad + Ours & \textbf{131}\dgood{$-$13\%} & \textbf{17.6}\dgood{+13\%} & \textbf{27.3}\dgood{+56\%} & \textbf{0.80}\dgood{+0.76} \\
\midrule
Wan2.1-14B~\cite{wan2025wan} & 181 & 16.7 & 18.2 & 0.05 \\
\rowcolor{gray!12}
\quad + Ours & \textbf{164}\dgood{$-$10\%} & \textbf{17.6}\dgood{+5\%} & \textbf{30.9}\dgood{+69\%} & \textbf{0.78}\dgood{+0.73} \\
\bottomrule
\end{tabular}
}
\end{minipage}
\end{table*}

% --------------------------------------------
\subsection{Ablation Study}
\label{sec:ablation}

% All ablations share the backbone, data, training budget and sampler of the full model, and each differs from it in a single component.
% We evaluate each claim on the test that targets it, and compare only variants that differ in the component under study. 

All ablations are conducted on thyroid data. Each variant shares the backbone, training data, training budget and sampler of the full model and differs from it in a single component. We use the metrics of Tab.~\ref{tab:baseline} and extend revisit consistency with a second reference: besides the real scan (GT), each revisited frame is also compared with the model's own earlier generation at the same pose (Self). GT measures whether revisited anatomy is correct, whereas Self measures whether the model remembers what it generated, independently of whether that was correct. Each claim is evaluated on the test that targets it, and only variants that differ in the component under study are compared.

% --------------------------------------------
% \paragraph{Beyond pose control: prior and memory.}
% The upper part of Tab.~\ref{tab:abl_mem_prior} starts from pose control alone and adds the two parts of the mental model.
% With pose control alone, the model barely follows the probe and produces the least realistic videos: the pose tells it where the probe is, but not what anatomy to expect there.
% The anatomical prior supplies this missing knowledge, and pose following and realism improve markedly.
% The memory addresses a different failure: it keeps revisited regions consistent with what was generated before, but on its own it does not make the anatomy correct, as the model then consistently repeats anatomy that it has invented.
% Only with both does the model generate plausible anatomy in unseen regions and consistent anatomy on revisits, and it is best on all metrics.

\paragraph{Beyond pose control: prior and memory.}
The upper part of Tab.~\ref{tab:abl_mem_prior} starts from pose control alone and adds the two proposed components.
With pose control alone, the model learns only a generic \emph{motion prior} of how scans typically evolve: it barely follows the probe and produces the least realistic videos, since the pose tells it where the probe is, but not what anatomy to expect there.
The anatomical prior supplies this missing knowledge, markedly improving pose following (Fol.\ 0.10 $\rightarrow$ 0.58) and generation realism (FID 40.2 $\rightarrow$ 30.7).
The memory instead lets the model remember: Self PSNR rises by about 10\,dB, so revisited regions reproduce what was generated before. Remembering, however, is not being correct. Without the prior, a thyroid that should appear beside the carotid artery as the probe moves forward (Fig.~\ref{fig:teaser}) may never be generated, and the memory then faithfully repeats this error on the way back, leaving GT PSNR unchanged.
Only with both, the prior for plausible anatomy and the memory for consistent revisits, does generation become anatomically consistent, with the best result on all metrics.

% --------------------------------------------
\paragraph{Fine-grained memory at the input, coarse prior across blocks.}
The memory provides fine-grained information for each token, whereas the prior is only a coarse population estimate of the layout. We therefore concatenate the memory with the input and add the prior to the early blocks (Sec.~\ref{sec:arch}).
The lower part of Tab.~\ref{tab:abl_mem_prior} keeps both conditions and changes only how they are injected.
When both are injected in the same form, either layer-wise or concatenated, combining them hurts rather than helps: FID becomes worse than with the memory alone (31.2) or the prior alone (30.7).
Sharing one pathway makes the two conditions compete, so the model cannot use both at full strength.
Dual injection removes this conflict. Each condition keeps its own pathway, and their benefits add up: it is the only variant that improves on both single-condition models, with the largest gain in pose following.

% --------------------------------------------
\paragraph{Model-agnostic gains from explicit anatomy.}
We add our prior and memory to three video diffusion backbones of different architecture and size: Wan2.2-5B, LTX2.5-22B and Wan2.1-14B (Tab.~\ref{tab:model_agnostic}).
Scaling up the backbone alone does not help: all three base models fail to follow the probe (Fol.\ $\leq$ 0.10) or recall revisited anatomy (Self $\leq$ 18.3\,dB), and the larger ones are no better than Wan2.2-5B.
With our conditioning, every backbone improves in realism, revisit consistency and, most clearly, pose following (Fol.\ 0.73--0.80).
This is consistent with the premise of this work: with scarce ultrasound data, a larger model does not learn the missing anatomical knowledge implicitly, whereas providing it explicitly improves every backbone.

% ---- Static and dynamic prior ----
% \paragraph{Refining the prior with every observation.}

\begin{figure}[b]
  \centering
  \includegraphics[width=\linewidth]{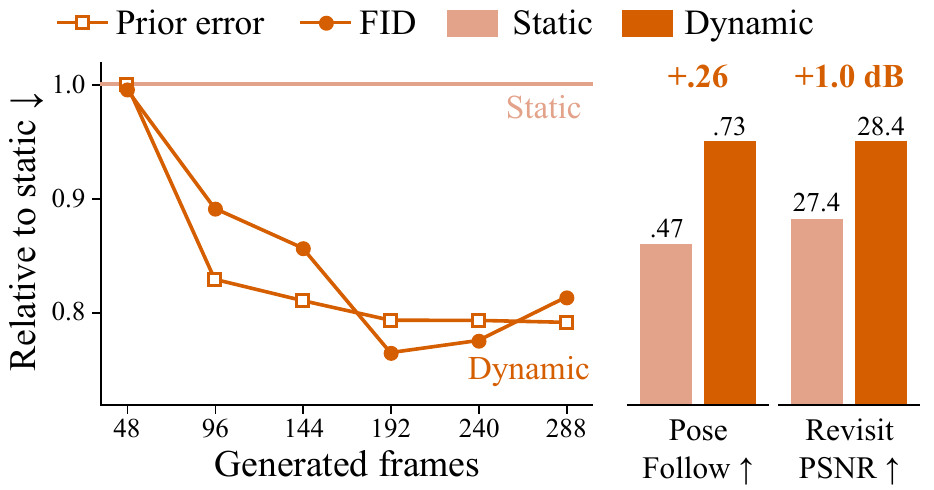}
  \caption{\textbf{Static vs.\ dynamic prior.} The dynamic prior lowers the prior error (distance to the real anatomy at unscanned planes) and the FID (left), and improves pose following and revisit PSNR (right).}
  \label{fig:prior_update}
\end{figure}

% --------------------------------------------
\begin{figure*}[t]
    \centering
    \includegraphics[width=\linewidth]{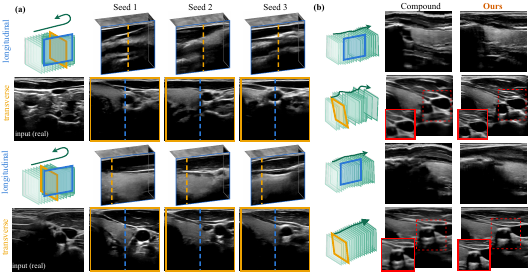}
    \caption{\textbf{Beyond recorded scans.} (a) Diverse scan synthesis: from one real frame and a planned path, each seed yields a different yet 3D-consistent anatomy. (b) Novel-plane synthesis from a real sweep: our model renders sharper planes than reslicing the compound.}
    \label{fig:app_qual_div}
\end{figure*}

\paragraph{Refining the prior with every observation.}
Like our expectation of an unfamiliar room, the prior should become specific to the subject as the scan proceeds. A static prior cannot: it conditions every chunk on the population mean $\boldsymbol{\mu}$, which carries little information about the specific subject. Our prior instead is updated after every chunk (Eq.~\ref{eq:prior_update}). \cref{fig:prior_update} compares the two on long thyroid scans. On the left, the prior error is the relative squared error between the prior resliced at each chunk and the real latents of that chunk. Before anything has been observed, the two priors coincide; once the first chunk has been generated, the error of the updated prior drops below that of the static one and stays lower. Generation performance follows the same trend, and the model with the updated prior degrades more slowly over the rollout (FID). On the right, we evaluate full rollouts with the protocols of Tab.~\ref{tab:abl_mem_prior}: the updated prior also improves pose following and revisit consistency.

% ---- Beyond Recorded Scans ----
\subsection{Qualitative Results: Beyond Recorded Scans}
\label{sec:app}
The model is not limited to reproducing recorded scans: from a single frame it generates diverse new scans, and from a real sweep it generates unscanned planes that were never acquired (Fig.~\ref{fig:app_qual_div}).

\paragraph{Anatomically Diverse Scan Synthesis}
\label{sec:app_plan}
Since the probe path is an input to the model, it can be freely defined rather than recorded; here we use a sweep followed by a turn. Given one real frame, the model generates the whole scan, and changing the seed changes the sampled anatomy while the probe pose stays the same. In Fig.~\ref{fig:app_qual_div}a, the three seeds produce thyroid lobes that differ in shape, echotexture and vessel layout, yet within each seed the transverse and longitudinal views agree at their intersection (dashed lines), \ie, each sample is a consistent 3D anatomy rather than a set of independent frames. Such diverse scans can enrich the scarce ultrasound training data with plausible anatomical variation, \eg, for training ultrasound world models.

\paragraph{Novel-Plane Synthesis from a Real Sweep}
\label{sec:app_rescan}
Freehand scanning is never steady, leaving gaps and tilts between neighboring slices, so the reslice at a new plane is blurred or smeared (Compound in Fig.~\ref{fig:app_qual_div}b).
Just as the probe path, the memory can also be supplied from outside rather than built from generated frames. We therefore encode the real sweep with the VAE, compound its latents into the memory (Eq.~\ref{eq:memory_update}) and update the prior from it, exactly as for generated chunks. The model then generates along a target path that moves the probe from the sweep to the new plane, \eg, by rotating it, conditioned on both the memory and the updated prior. It recovers sharp texture and fine structures lost in the compound (zoom-ins), correcting compounding errors and yielding more realistic images than reslicing alone.
\section{Conclusion}
\label{sec:conclusion}
Video generation models anticipate unseen content from priors learned implicitly from large-scale data. For anatomically consistent ultrasound video generation, this is insufficient: probe pose tells the model where to scan, but not what anatomy to expect, and scarce ultrasound data prevent learning this implicitly. We provide the generator with two parts of a sonographer's mental model: an anatomical prior, refined online by each generated chunk, for unscanned regions, and a 3D latent memory built by compounding for scanned regionspreviously generated frames for scanned regions. Together, they keep the generated anatomy plausible where nothing has been scanned yet and consistent where it has been scanned before. Our evaluation focuses on generation fidelity and anatomical consistency, leaving its effectiveness for embodied planning, where generated views guide the probe motion, untested. In future work, we plan to use it as a world model for robotic ultrasound, \eg, to train and evaluate probe-navigation policies or plan probe motions by predicting candidate views and selecting the path that best reaches a target view~\cite{assran2025v}.

{
    \small
    \bibliographystyle{ieeenat_fullname}
    \bibliography{main}
}

% Appendix / supplementary material: not included in the arXiv version
% \input{sec/X_suppl}

\end{document}